\documentclass{article} 
\usepackage{iclr2015,times}
\usepackage{url}
\usepackage{bbm}
\usepackage{floatrow}
\newfloatcommand{btabbox}{table}[][\FBwidth]
\usepackage{times}
\usepackage{epsfig}
\usepackage{graphicx}
\usepackage{amsmath}
\usepackage{amssymb}
\usepackage{array}
\usepackage{dsfont}
\usepackage{multirow}
\usepackage{subfigure}
\usepackage{color}
\definecolor{gray}{rgb}{0.5,0.5,0.5}
\iftrue 
	\newcommand{\topic}[1]{\textcolor{gray}{\textbf{(#1.)}}}
	\newcommand{\outline}[1]{{\textcolor{blue}{[[{#1}]]}}}
	\newcommand{\commenttext}[1]{\textcolor{red}{[[{#1}]]}}
	\newcommand{\commentfoot}[1]{\footnote{\textcolor{red}{\textit{#1}}}}
\else 
	\newcommand{\topic}[1]{}
	\newcommand{\outline}[1]{}
	\newcommand{\commenttext}[1]{}
	\newcommand{\commentfoot}[1]{}
\fi

\usepackage[pagebackref=true,breaklinks=true,letterpaper=true,colorlinks,bookmarks=false]{hyperref}

\usepackage{algpseudocode}
\usepackage{algorithm}

\graphicspath{{./images/}}

\def\1{\mathds{1}}

\makeatletter
\DeclareRobustCommand\onedot{\futurelet\@let@token\@onedot}
\def\@onedot{\ifx\@let@token.\else.\null\fi\xspace}

\makeatother

\usepackage{framed}
\usepackage{xcolor}
\definecolor{gainsboro}{RGB}{220,220,220}

\iclrfinalcopy 

\begin{document}

\title{Training deep neural networks\\ on noisy labels with bootstrapping}
\author{
Scott E.~Reed \& Honglak Lee  \\
Dept. of Electrical Engineering and Computer Science, University of Michigan\\
Ann Arbor, MI, USA \\
\texttt{\{reedscot,honglak\}@umich.edu} \\
\And
Dragomir Anguelov, Christian Szegedy, Dumitru Erhan \& Andrew Rabinovich \\
Google, Inc.\\
Mountain View, CA, USA \\
\texttt{\{dragomir,szegedy,dumitru,amrabino\}@google.com} \\
}

\maketitle

\begin{abstract}
Current state-of-the-art deep learning systems for visual object recognition and detection use purely supervised training with regularization such as dropout to avoid overfitting.
The performance depends critically on the amount of labeled examples, and in current practice the labels are assumed to be unambiguous and accurate.
However, this assumption often does not hold; e.g. in recognition, class labels may be missing; in detection, objects in the image may not be localized; and in general, the labeling may be subjective.
In this work we propose a generic way to handle noisy and incomplete labeling by augmenting the prediction objective with a notion of consistency.
We consider a prediction \emph{consistent} if the same prediction is made given similar percepts, where the notion of similarity is between deep network features computed from the input data.
In experiments we demonstrate that our approach yields substantial robustness to label noise on several datasets.
On MNIST handwritten digits, we show that our model is robust to label corruption.
On the Toronto Face Database, we show that our model handles well the case of subjective labels in emotion recognition, achieving state-of-the-art results, and can also benefit from unlabeled face images with no modification to our method.
On the ILSVRC2014 detection challenge data, we show that our approach extends to very deep networks, high resolution images and structured outputs, and results in improved scalable detection.
\end{abstract}
\section{Introduction}
Currently the predominant systems for visual object recognition and detection~\citep{krizhevsky2012imagenet,zeiler2013visualizing,girshick2013rich,sermanet2013overfeat,szegedy2014scalable} use purely supervised training with regularization such as dropout~\citep{hinton2012improving} to avoid overfitting.
These systems do not account for missing labels, subjective labeling or inexhaustively-annotated images.
However, this assumption often does not hold, especially for very large datasets and in high-resolution images with complex scenes.
For example, in recognition, the class labels may be missing; in detection, the objects in the image may not all be localized; in subjective tasks such as facial emotion recognition, humans may not even agree on the class label.
As training sets for deep networks become larger (as they should), the problem of missing and noisy labels becomes more acute, and so we argue it is a fundamental problem for scaling up vision.

In this work we propose a simple approach to hande noisy and incomplete labeling in weakly-supervised deep learning, by augmenting the usual prediction objective with a notion of perceptual consistency.
We consider a prediction \emph{consistent} if the same prediction is made given similar percepts, where the notion of similarity incorporates features learned by the deep network.

One interpretation of the perceptual consistency objective is that the learner makes use of its representation of the world (implicit in the network parameters) to match incoming percepts to known categories, or in general structured outputs.
This provides the learner justification to ``disagree'' with a perceptually-inconsistent training label, and effectively re-label the data while training.
More accurate labels may lead to a better model, which allows further label clean-up, and the learner bootstraps itself in this way.
Of course, too much skepticism of the labels carries the risk of ending up with a delusional agent, so it is important to balance the trade-off between prediction and the learner's perceptual consistency.

In our experiments we demonstrate that our approach yields substantial robustness to several types of label noise on several datasets.
On MNIST handwritten digits~\citep{lecun1998mnist} we show that our model is robust to label corruption.
On the Toronto Face Database~\citep{susskind2010toronto} we show that our model handles well the case of subjective labels in emotion recognition, achieving state-of-the-art results, and can also benefit from unlabeled face images with no modification to our method.
On the ILSVRC2014 detection challenge data~\citep{ILSVRCarxiv14}, we show that our approach 
improves single-shot person detection using a MultiBox network~\citep{erhan14scalable}, and also improves performance in full 200-way detection using MultiBox for region proposal and a deep CNN for post-classification.

In section~\ref{sec:related} we discuss related work, in section~\ref{sec:method} we describe our method along with a probabilistic interpretation and in section~\ref{sec:experiments} we present our results.
\section{Related Work}
\label{sec:related}
The literature on semi-supervised and weakly-supervised learning is vast (see~\citet{zhu2005semi} for a survey), and so in this section we focus on the key previous papers that inspired this work and on other papers on weakly- and semi-supervised deep learning.

The notion of bootstrapping, or ``self-training'' a learning agent was proposed in~\citep{yarowsky1995unsupervised} as a way to do word-sense disambiguation with only unlabeled examples and a small list of seed example sentences with labels. The algorithm proceeds by building an initial classifier using the seed examples, and then iteratively classifying unlabeled examples, extracting new seed rules for the classifier using the now expanded training data, and repeating these steps until convergence. The algorithm was analyzed by Abney~\citep{abney2004understanding} and more recently by~\citep{haffari2012analysis}.

Co-training~\citep{blum1998combining,nigam2000text,nigam2000analyzing} was similarly-motivated but used a pair of classifiers with separate views of the data to iteratively learn and generate additional training labels.
\citet{whitney2012bootstrapping} proposed bootstrapping labeled training examples with graph-based label propagation. \citet{brodley1999identifying} developed statistical methods for identifying mislabeled training data.

\citet{rosenberg2005semi} also trained an object detection system in a weakly-supervised manner using self-training, and demonstrated that their proposed model achieved comparable performance to models trained with a much larger set of labels. However, that approach works as a wrapper around an existing detection system, whereas in this work we integrate a consistency objective for bootstrapping into the training of the deep network itself.

Our work shares a similar motivation to these earlier works, but instead of explicitly generating new training labels and adding new examples to the training set in an outer loop, we incorporate our consistency objective directly into the model. In addition, we consider not only the case of learning from unlabeled examples, but also from noisy labels and inexhaustively-annotated examples.

\citet{mnih2012learning} developed deep neural networks for improved labeling of aerial images, with robust loss functions to handle label omission and registration errors. This work shares a similar motivation of robustness to noisy labels, but rather than formulating loss functions for specific types of noise, we add a generic consistency objective to the loss to achieve robustness.

Minimum entropy regularization, proposed in~\citep{grandvalet2005semi,grandvalet2006}, performs semi-supervised learning by augmenting cross-entropy loss with a term encouraging the classifier to make predictions with high confidence on the unlabeled examples\footnote{see eq. 9.7 in ~\citep{grandvalet2006}}. This is notable because in their approach training on unlabeled examples does not require a generative model, which is beneficial for training on high-resolution images and other sensory data.
We take a similar approach by side-stepping the difficulty of fully-generative models of high-dimensional sensory data. However, we extend beyond shallow models to deep networks, and to structured output prediction.

Never ending language learning (NELL)~\citep{carlson2010toward} and never ending image learning (NEIL)~\citep{chen2013neil,chen2014enriching} are lifelong-learning systems for language and image understanding, respectively. They continuously bootstrap themselves using a cycle of data collection, propagation of labels to the newly collected data, and self-improvement by training on the new data.
Our work is complementary to these efforts, and focuses on building robustness to noisy and missing labels into the model for weakly-supervised deep learning.

\citet{larochelle2008classification} developed an RBM for classification that uses a hybrid generative and discriminative training objective. Deep Boltmann Machines~\citep{salakhutdinov2009deep} can also be trained in a semi-supervised manner with labels connected to the top layer. More recently, multi-prediction DBM training~\citep{goodfellow2013multi} and Generative Stochastic Networks~\citep{bengio2013deep} improved the performance and simplified the training of deep generative models, enabling training via backpropagation much like in standard deep supervised networks. 
%
However, fully-generative unsupervised training on high-dimensional sensory data, e.g. ImageNet images, is still far behind supervised methods in terms of performance, and so in this work we do not follow the generative approach directly.
Instead, this work focuses on a way to benefit from unlabeled and weakly-labeled examples with minimal modification to existing deep supervised networks. We demonstrate increased robustness to label noise and performance improvements from unlabeled data for a minimal engineering effort.

More recently, the problem of deep learning from noisy labels has begun to receive attention. \citet{lee2013pseudo} also followed the idea of minimum entropy regularization, and proposed generating ``pseudo-labels'' as training targets for unlabeled data, and showed improved performance on MNIST with few labeled examples.  \citet{sukhbaatar2014learning} developed two deep learning techniques for handling noisy labels, learning to model the noise distribution in a top-down and bottom-up fashion. 
In this work, we push further by extending beyond class labels to structured outputs, and we achieve state-of-the-art scalable detection performance on ILSVRC2014, despite the fact that our method does not require explicitly modeling the noise distribution.
\section{Method}
\label{sec:method}
In this section we describe two approaches: section~\ref{sec:multiclass} uses reconstruction error as a consistency objective and explicitly models the noise distribution as a matrix mapping model predictions to training labels.
A reconstruction loss is added to promote top-down consistency of model predictions with the observations, which allows the model to discover the pattern of noise in the data.
%
%

The method presented in section~\ref{sec:bootstrap} (bootstrapping) uses a convex combination of training labels and the \emph{current} model's predictions to generate the training targets, and thereby avoids directly modeling the noise distribution.
This property is well-suited to the case of structured outputs, for which modeling dense interactions among all pairs of output units may be neither practical nor useful.
%
%
These two approaches are compared empirically in section~\ref{sec:experiments}.

In section~\ref{sec:structured} we show how to apply our bootstrapping approach to structured outputs by using the MultiBox~\citep{erhan14scalable} region proposal network to handle the case of inexhaustive structured output labeling for single-shot person detection and for class-agnostic region proposal.
%
%
\subsection{Consistency in multi-class prediction via reconstruction}
\label{sec:multiclass}
Let $\mathbf{x} \in \{0,1\}^{D}$ be the data (or deep features computed from the data) and $\mathbf{t} \in \{0,1\}^{L}, \sum_{k}t_{k} = 1$ the observed noisy multinomial labels.
The standard softmax regresses $\mathbf{x}$ onto $\mathbf{t}$ without taking into account noisy or missing labels.
In addition to optimizing the conditional log-likelihood $\log P(\mathbf{t} | \mathbf{x})$, we add a regularization term encouraging the class prediction to be perceptually consistent.

We first introduce into our model the ``true'' class label (as opposed to the noisy label observations) as a latent multinomial variable $\mathbf{q} \in \{0,1\}^{L}, \sum_{j}q_{j} = 1$.
Our deep feed-forward network models the posterior over $\mathbf{q}$ using the usual softmax regression:
\begin{align}
P(q_{j} = 1 | \mathbf{x}) = \dfrac{\tilde{P}(q_{j} = 1 | \mathbf{x})}{\sum_{j'=1}^{L}\tilde{P}(q_{j'} = 1 | \mathbf{x})}=
\dfrac{\exp(\sum_{i=1}^{D}W^{(1)}_{ij}x_{i} + b^{(1)}_{i})}{\sum_{j'=1}^{L}\exp(\sum_{i=1}^{D}W^{(1)}_{ij'}x_{i}+b^{(1)}_{i})}
\label{eq:prediction}
\end{align}
where $\tilde{P}$ denotes the unnormalized probability distribution.
Given the true label $\mathbf{q}$, the label noise can be modeled using another softmax with logits as follows:
\begin{align}
\log \tilde{P}(t_{k}=1|\mathbf{q}) = \sum_{j=1}^{L}W^{(2)}_{kj}q_{j} + b^{(2)}_{k}
\end{align}
%
%
Roughly, $W^{(2)}_{kj}$ learns the log-probability of observing true label $j$ as noisy label $k$.
Given only an observation $\mathbf{x}$, we can marginalize over $\mathbf{q}$ to compute the posterior of target $\mathbf{t}$ given $\mathbf{x}$.
\begin{align}
P(t_{k} = 1 | \mathbf{x}) &= \sum_{j=1}^{L}P(t_{k} = 1, q_{j} = 1 | \mathbf{x})
= \sum_{j=1}^{L}P(t_{k} = 1 | q_{j} = 1)P(q_{j} = 1 | \mathbf{x})
\label{eq:softmax_cons}
\end{align}
%
where the label noise distribution and posterior over true labels are defined above. 
We can perform discriminative training by gradient ascent on $\log P(\mathbf{t} | \mathbf{x})$.
However, this purely discriminative training does not yet incorporate perceptual consistency, and there is no explicit incentive for the model to treat $\mathbf{q}$ as the ``true'' label; it can be viewed as another hidden layer, with the multinomial constraint resulting in an information bottleneck.

In unpublished work\footnote{Known from personal correspondence.}, \citet{hinton2009rbm} developed a Restricted Boltzmann Machine (RBM)~\citep{smolensky1986information} variant with hidden multinomial output unit  $\mathbf{q}$ and observed noisy label unit $\mathbf{t}$ as described above. The associated energy function can be written as
\begin{align}
E(\mathbf{x},\mathbf{t},\mathbf{q}) = -\sum_{i=1}^{D}\sum_{j=1}^{L}W^{(1)}_{ij}x_{i}q_{j} - \sum_{k=1}^{L}\sum_{j=1}^{L}W^{(2)}_{kj}t_{k}q_{j} - \sum_{j=1}^{L}b^{(1)}_{j}q_{j} -\sum_{k=1}^{L}b^{(2)}_{k}t_{k}-\sum_{i=1}^{D}b^{(3)}_{i}x_{i} 
\label{eq:energy}
\end{align}
Due to the bipartite structure of the RBM, $\mathbf{t}$ and $\mathbf{x}$ are conditionally independent given $\mathbf{q}$, and so the energy function in eq.~\eqref{eq:energy} leads to a similar form of the posterior as in eq.~\eqref{eq:softmax_cons}, marginalizing out the hidden multinomial unit. The probability distribution arising from~\eqref{eq:energy} is given by
\begin{align*}
P(\mathbf{x},\mathbf{t}) = \dfrac{\sum_{\mathbf{q}}\exp(-E(\mathbf{x},\mathbf{t},\mathbf{q}))}{Z}
\end{align*}
where $Z = \sum_{\mathbf{x},\mathbf{t},\mathbf{q}}\exp(-E(\mathbf{x},\mathbf{t},\mathbf{q}))$ is the partition function. The model can be trained with a generative objective, e.g. by approximate gradient ascent on $\log P(\mathbf{x},\mathbf{t})$ via contrastive divergence~\citep{hinton2002training}.
Generative training naturally provides a notion of consistency between observations $\mathbf{x}$ and predictions $\mathbf{q}$ because the model learns to draw sample observations via the conditional likelihood $P(x_{i} = 1 | \mathbf{q}) = \sigma(\sum_{j=1}^{L}W^{(1)}_{ij}q_{j} + b^{(3)}_{i})$, assuming binary observations.
\begin{figure}[h]
\centering
   \includegraphics[width =0.7\textwidth] {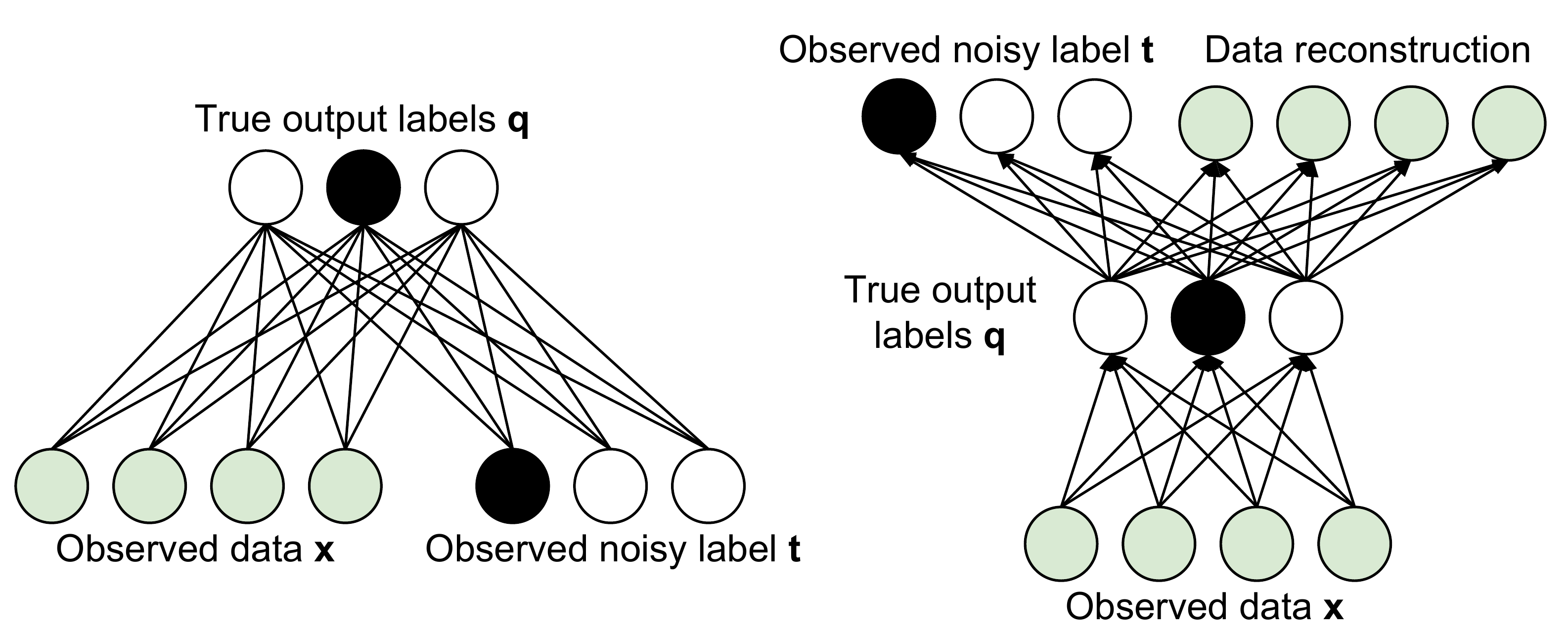}
   \caption{Left: Restricted Boltzmann Machine with hidden multinomial output unit. Right: Analogous feed-forward autoencoder version.}
   \label{fig:rbm_ae_multiclass}
\end{figure}

However, fully-generative training is complicated by the fact that the exact likelihood gradient is intractable due to computing the partition function $Z$, and in practice MCMC is used. Generative training is further complicated (though certainly still possible) in cases where the features $\mathbf{x}$ are non-binary. To avoid these complications, and to make our approach rapidly applicable to existing deep networks using rectified linear activations, and trainable via exact gradient descent, we propose an analogous autoencoder version.

Figure~\ref{fig:rbm_ae_multiclass} compares the RBM and autoencoder approaches to the multiclass prediction problem with perceptual consistency. The overall objective in the feed-forward version is as follows:
\begin{align}
\mathcal{L}_{recon}(\mathbf{x},\mathbf{t}) = -\sum_{k=1}^{L} t_{k} \log P(t_{k}=1 | \mathbf{x}) + \beta|| \mathbf{x} - W^{(2)}\mathbf{q}(\mathbf{x})||^{2}_{2}
\label{eq:cons_ff}
\end{align}
where $\mathbf{q}(\mathbf{x})_{j} = P(q_{j} = 1 | \mathbf{x})$ as in equation~\ref{eq:prediction}.
The parameter $\beta$ can be found via cross-validation.
Experimental results using this method are presented in sections~\ref{sec:mnist} and~\ref{sec:tfd}.
\subsection{Consistency in multi-class prediction via bootstrapping}
\label{sec:bootstrap}
In this section we develop a simple consistency objective that does not require an explicit noise distribution or a reconstruction term. The idea is to dynamically update the targets of the prediction objective based on the current state of the model. The resulting targets are a convex combination of (1) the noisy training label, and (2) the current prediction of the model.
Intuitively, as the learner improves over time, its predictions can be trusted more.
This mitigates the damage of incorrect labeling, because incorrect labels are likely to be eventually highly inconsistent with other stimuli predicted to have the same label by the model.

By paying less heed to inconsistent labels, the learner can develop a more coherent model, which further improves its ability to evaluate the consistency of noisy labels.
We refer to this approach as ``bootstrapping'', in the sense of pulling oneself up by one's own bootstraps, and also due to inspiration from the work of~\citet{yarowsky1995unsupervised} which is also referred to as bootstrapping. 

Concretely, we use a cross-entropy objective as before, but generate new regression targets for each SGD mini-batch based on the current state of the model.
We empirically evaluated two types of bootstrapping. ``Soft'' bootstrapping uses predicted class probabilities $\mathbf{q}$ directly to generate regression targets for each batch as follows:
\begin{align}
\mathcal{L}_{soft}(\mathbf{q},\mathbf{t}) &= \sum_{k=1}^{L}[\beta t_{k} + (1-\beta) q_{k}]\log(q_{k})
\end{align}
In fact, it can be shown that the resulting objective is equivalent to softmax regression with minimum entropy regularization, which was previously studied in~\citep{grandvalet2006}. Intuitively, minimum entropy regularization encourages the model to have a high confidence in predicting labels (even for the unlabeled examples, which enables semi-supervised learning). 

``Hard'' bootstrapping modifies regression targets using the MAP estimate of $\mathbf{q}$ given $\mathbf{x}$,
 which we denote as $z_{k} :=  \mathbbm{1}[k = \text{argmax } q_{i}, i = 1...L]$:
\begin{align}
\mathcal{L}_{hard}(\mathbf{q},\mathbf{t}) &= \sum_{k=1}^{L}[\beta t_{k} + (1-\beta) z_{k}]\log(q_{k})
\label{eq:softmax_bootstrap_hard}
\end{align}
When used with mini-batch stochastic gradient descent, this leads to an EM-like algorithm: In the E-step, estimate the ``true'' confidence targets as a convex combination of training labels and model predictions; in the M-step, update the model parameters to better predict those generated targets.

Both hard and soft bootstrapping can be viewed as instances of a more general approach in which model-generated regression targets are modulated by a softmax temperature parameter $T$; i.e.
\begin{align}
P(q_{j}=1 | \mathbf{x}) = \dfrac{\exp (T \cdot( \sum_{i=1}^{D}W^{(1)}_{ij}x_{i} + b^{(1)}_{j}))}{\sum_{j'=1}^{L}\exp (T \cdot( \sum_{i=1}^{D}W^{(1)}_{ij'}x_{i} +b^{(1)}_{j'}))}
\end{align}
Setting $T=1$ recovers soft boostrapping, and $T = \infty$ recovers hard bootstrapping. We only use these two operating points in our experiments, but it may be worthwhile to explore other values for $T$, and learning $T$ for each dataset.
\subsection{Consistency with structured output prediction}
\label{sec:structured}
Noisy labels also occur in structured output prediction problems such as object detection.
Current state-of-the-art object detection systems train on images annotated with bounding box labels of the relevant objects in each image, and the class label for each box.
However, it is expensive to exhaustively annotate each image, and for some commonly-appearing categories the data may be prone to missing annotations.
In this section, we modify the training objective of the MultiBox~\citep{erhan14scalable} network for object detection to incorporate a notion of perceptual consistency into the loss.

In the MultiBox approach, ground-truth bounding boxes are clustered and the resulting centroids are used as ``priors'' for predicting object location.
A deep neural network is trained to predict, for each groundtruth object in an image, a residual of that groundtruth bounding box to the best-matching bounding box prior.
The network also outputs a logistic confidence score for each prior, indicating the model's belief of whether or not an object appears in the corresponding location.
Because MultiBox gives proposals with confidence scores, it enables very efficient runtime-quality tradeoffs for detection via thresholding the top-scoring proposals within budget.
Thus it is an attractive target for further quality improvements, as we pursue in this section.

Denote the confidence score training targets as $\mathbf{t} \in \{0,1\}^{L}$ 
and the predicted confidence scores as 
$\mathbf{c} \in [0,1]^{L}$. 
The objective for MultiBox~\footnote{We omit the bounding box regression term for simplicity, see~\citep{erhan14scalable} for full details.} can be written as the following cross-entropy loss:
\begin{align}
\mathcal{L}_{multibox}(\mathbf{c},\mathbf{t}) &= -\sum_{k=1}^{L}( t_{k} \log (c_{k}) + (1 - t_{k})\log(1 - c_{k}))
\label{eq:multibox}
\end{align}
Note that the sum here is over object locations, not class labels as in the case of sections~\ref{sec:multiclass} and~\ref{sec:bootstrap}.

If there is an object at location $k$, but $t_{k} = 0$ due to inexhaustive annotation, the model pays a large cost for correctly predicting $c_{k} = 1$.
Training naively on the noisy labels leads to perverse learning situations such as the following: two objects of the same category (potentially within the same image) appear in the training data, but only one of them is labeled.
To reduce the loss, the confidence prediction layer must learn to distinguish the two objects, which is exactly contrary to the objective of visual invariance to category-preserving differences.

To incorporate a notion of perceptual consistency into the loss, 
we follow the same approach as in the case of multi-class classification: augment the regression targets using the model's current state. In the ``hard'' case, MAP estimates can be obtained by thresholding $c_{k} > 1/2$.
\begin{align}
\label{eq:xent_hard}
\mathcal{L}_{multibox-hard}(\mathbf{c},\mathbf{t}) =& -\sum_{k=1}^{L}[\beta t_{k} + (1-\beta) \mathbbm{1}_{c_{k}>0.5}] \log (c_{k}) \\
& -\sum_{k=1}^{L} [\beta(1 - t_{k}) + (1-\beta)(1 - \mathbbm{1}_{c_{k} > 0.5})]\log(1 - c_{k})\nonumber
\end{align}
\begin{align}
\label{eq:xent_soft}
\mathcal{L}_{multibox-soft}(\mathbf{c},\mathbf{t}) =& -\sum_{k=1}^{L}[\beta t_{k} + (1-\beta) c_{k}] \log (c_{k}) \\
& -\sum_{k=1}^{L} [\beta(1 - t_{k}) + (1-\beta)(1 - c_{k})]\log(1 - c_{k})\nonumber
\end{align}
With the bootstrap variants of the MultiBox objective, unlabeled positives pose less of a problem because penalties for large $c_{k}$ are down-scaled by factor $\beta$ in the first term and $(1-c_{k})$ in the second term.
By mitigating penalties due to missing positives in the data, our approach allows the model to learn to predict $c_{k}$ with high confidence even if the objects at location $k$ are often unlabeled.
\section{Experiments}
We perform experiments on three image understanding tasks: MNIST handwritten digits recognition, Toroto Faces Database facial emotion recognition, and ILSVRC2014 detection. In all tasks, we train a deep neural network with our proposed consistency objective. In our figures, ``bootstrap-recon'' refers to training as described in section~\ref{sec:multiclass}, using reconstruction as a consistency objective. ``bootstrap-soft'' and ``bootstrap-hard'' refer to our method described in sections~\ref{sec:bootstrap} and~\ref{sec:structured}.
\label{sec:experiments}
\subsection{MNIST with noisy labels}
\label{sec:mnist}
\begin{figure}[h]
\centering
\includegraphics[width=0.95\textwidth]{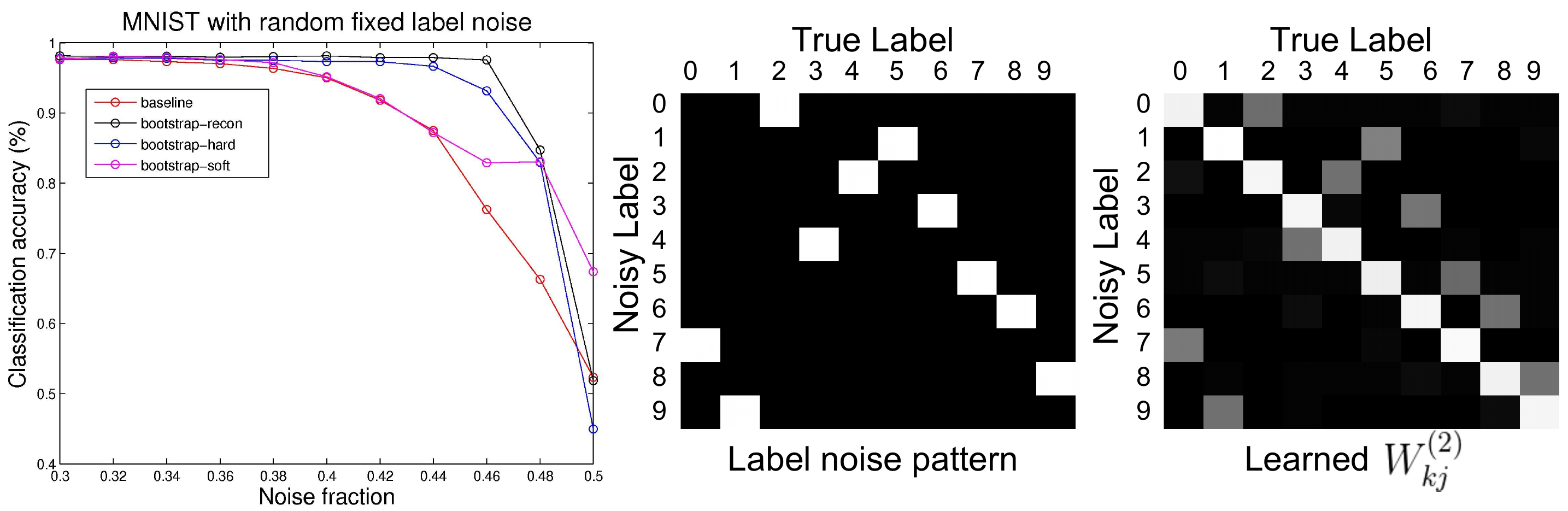}
\caption{Left: Digit recognition accuracy versus percent corrupted labels. Middle: a visualization of the noise pattern. If there is a white entry in row $r$, column $c$, then label $c$ is mapped to to label $r$ with some probability during training. Right: a visualization of $P(t_{r}=1 | q_{c} = 1)$ learned by our bootstrap-recon model (as parameters $W^{(2)}_{rc}$) trained with 40\% label noise.}
\label{fig:mnist_exp}
\end{figure}

In this section we train using our reconstruction-based objective (detailed in section~\ref{sec:multiclass}) on MNIST handwritten digits with varying degrees of noise in the labels.
Specifically, we used a fixed random permutation of the labels as visualized in figure~\ref{fig:mnist_exp}, and we perform control experiments while varying the probability of applying the label permutation to each training example.

All models were trained with mini-batch SGD, with the same architecture: 784-500-300-10 neural network with rectified linear units. We used $L_{2}$ weight decay of $0.0001$. We found that $\beta = 0.8$ worked best for bootstrap-hard, $0.95$ for bootstrap-soft, and $0.005$ for bootstrap-recon. We initialize $W^{(2)}$ to the identity matrix.

For the network trained with our proposed consistency objective, we initialized the network layers from the baseline prediction-only model. It is also possible to initialize from scratch using our approach, but we found that with pre-training we could use a larger $\beta$ and more quickly converge to a good result. Intuitively, this is similar to the initial collection of ``seed'' rules in the original bootstrapping algorithm of~\citep{yarowsky1995unsupervised}. During the fine-tuning training phase, all network weights are updated by backpropagating gradients through the layers.


Figure~\ref{fig:mnist_exp} shows that our bootstrapping method provides a very significant benefit in the case of permuted labels. The bootstrap-recon method performs the best, and bootstrap-hard nearly as well. The bootstrap-soft method provides some benefit in the high-noise regime, but only slightly better than the baseline overall.

Figure~\ref{fig:mnist_exp} also shows that bootstrap-recon effectively learns the noise distribution $P(\mathbf{t} | \mathbf{q	})$ via the parameters $W^{(2)}_{kj}$. 
Intuitively, the loss from the reconstruction term provides the learner a basis on which predictions $\mathbf{q}$ may disagree with training labels $\mathbf{t}$.
Since $\mathbf{x}$ must be able to be reconstructed from $\mathbf{q}$ in bootstrap-recon, learning a non-identity $W^{(2)}$ allows $\mathbf{q}$ to flexibly vary from $\mathbf{t}$ to better reconstruct $\mathbf{x}$ without incurring a penalty from prediction error.

However, it is interesting to note that bootstrap-hard achieves nearly equivalent performance without explicitly parameterizing the noise distribution. This is useful because reconstruction may be challenging in many cases, such as when $\mathbf{x}$ is drawn from a complicated, high-dimensional distribution, and bootstrap-hard is trivial to implement on top of existing deep supervised networks.
\subsection{Toronto Faces Database emotion recognition}
\label{sec:tfd}
In this section we present results on emotion recognition. The Toronto Faces Database has 112,234 images, 4,178 of which have emotion labels. In all experiments we first extracted spatial-pyramid-pooled OMP-1 features as described in~\citep{coates2011importance} to get $3200$-dimensional features. We then trained a 3200-1000-500-7 network to predict the 1-of-7 emotion labels for each image.

As in the case for our MNIST experiments, we initialize our model from the network pre-trained with prediction only, and the fine-tuned all layers with our hybrid objective.

\begin{figure}[h]
\CenterFloatBoxes
\begin{floatrow}
\ffigbox{%
  \includegraphics[width =0.4\textwidth] {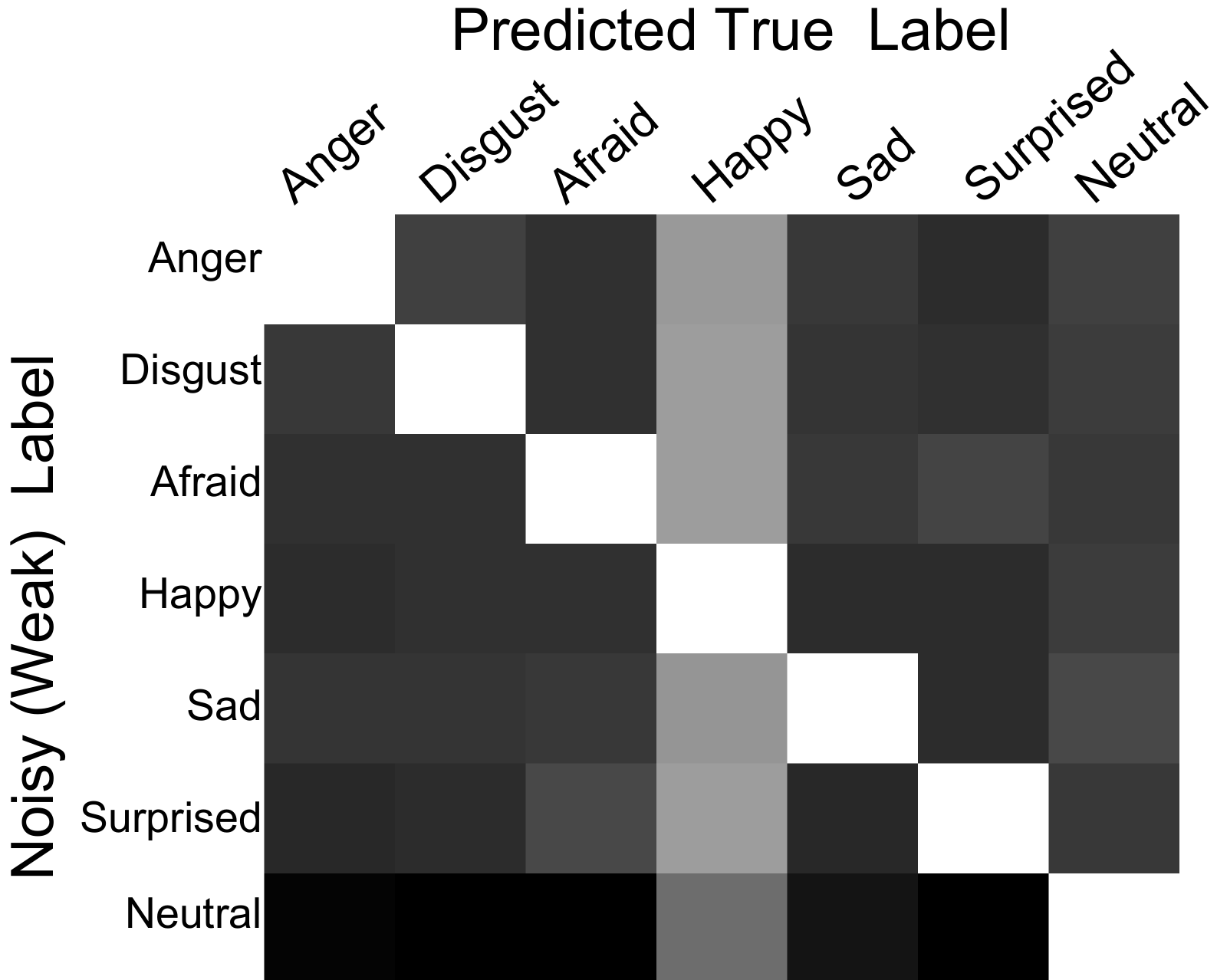}
} {%
  \caption{Predicted-to-noisy emotion label connection $W^{(2)}$ learned by our model.}%
\label{fig:tfd_exp}
}
\killfloatstyle
\btabbox{
  \begin{tabular}{| l | c |}
  \hline
  \textbf{Training} & \textbf{Accuracy (\%)}  \\
  \hline
  baseline & $85.3$ \\
  \hline
  bootstrap-recon & $\mathbf{86.8}$ \\
  \hline
  bootstrap-hard & $\mathbf{86.8}$ \\
  \hline
  bootstrap-soft & $85.6$ \\
  \hline
  disBM~\footnote{\citep{icml2014disentangling}} & $85.4$ \\
  \hline
  CDA+CCA~\footnote{\citep{rifai2012disentangling}} & $85.0$ \\
  \hline
  \end{tabular}
}{%
  \caption{Emotion recognition results on Toronto Faces Database compared to state-of-the-art methods.}%
}
\end{floatrow}
\end{figure}

Figure~\ref{fig:tfd_exp} summarizes our TFD results. As in the case of MNIST, bootstrap-recon and bootstrap-hard perform the best, significantly outperforming the softmax baseline, and bootstrap-soft provides a more modest improvement.

%
There is significant off-diagonal weight in $W^{(2)}$ learned by bootstrap-recon on TFD, suggesting that the model learns to ``hedge'' its emotion prediction during training by spreading probability mass from the predicted class to commonly-confused classes, such as ``afraid'' and ``surprised''. The strongest off diagonals are in the ``happy'' column, which may be due to the fact that ``happy'' is the most common expression, or perhaps that happy expressions have a large visual diversity. Our method improves the emotion recognition performance, which to our knowledge is state-of-the-art.
The performance improvement from all three bootstrap methods on TFD suggests that our approach can be useful not just for mistaken labels, but also for semi-supervised learning (missing labels) and learning from weak labels such as emotion categories.
\subsection{ILSVRC 2014 fast single-shot person detection}
\label{sec:ilsvrc}
In this section we apply our method to detecting persons using a MultiBox network built on top of the Inception architecture proposed in~\citep{szegedy2014going}. We first pre-trained MultiBox on class-agnostic localization using the full ILSVRC2014 training set, since there are only several thousand images labeled with persons in ILSVRC, and then fine-tuned on person images only.

An important point for comparison is the top-$K$ bootstrapping heuristic introduced for MultiBox training in~\citep{szegedy2014scalable} for person detection in the presence of missing annotations.
In that approach, the top-$K$ largest confidence predictions are dropped from the loss (which we show here in eq.~\eqref{eq:multibox}), and the setting used was $K=4$. In other words, there is no gradient coming from the top-$K$ most confident location predictions.
In fact, it can be viewed as a form of bootstrapping where only the top-$K$ most confident locations modify their targets, which become the predictions themselves.
%
In this work, we aim to achieve similar or better performance in a more general way that can be applied to MultiBox and other discriminative models.

\begin{figure}[h]
 \CenterFloatBoxes
\begin{floatrow}
\ffigbox{%
  \includegraphics[width =0.4\textwidth] {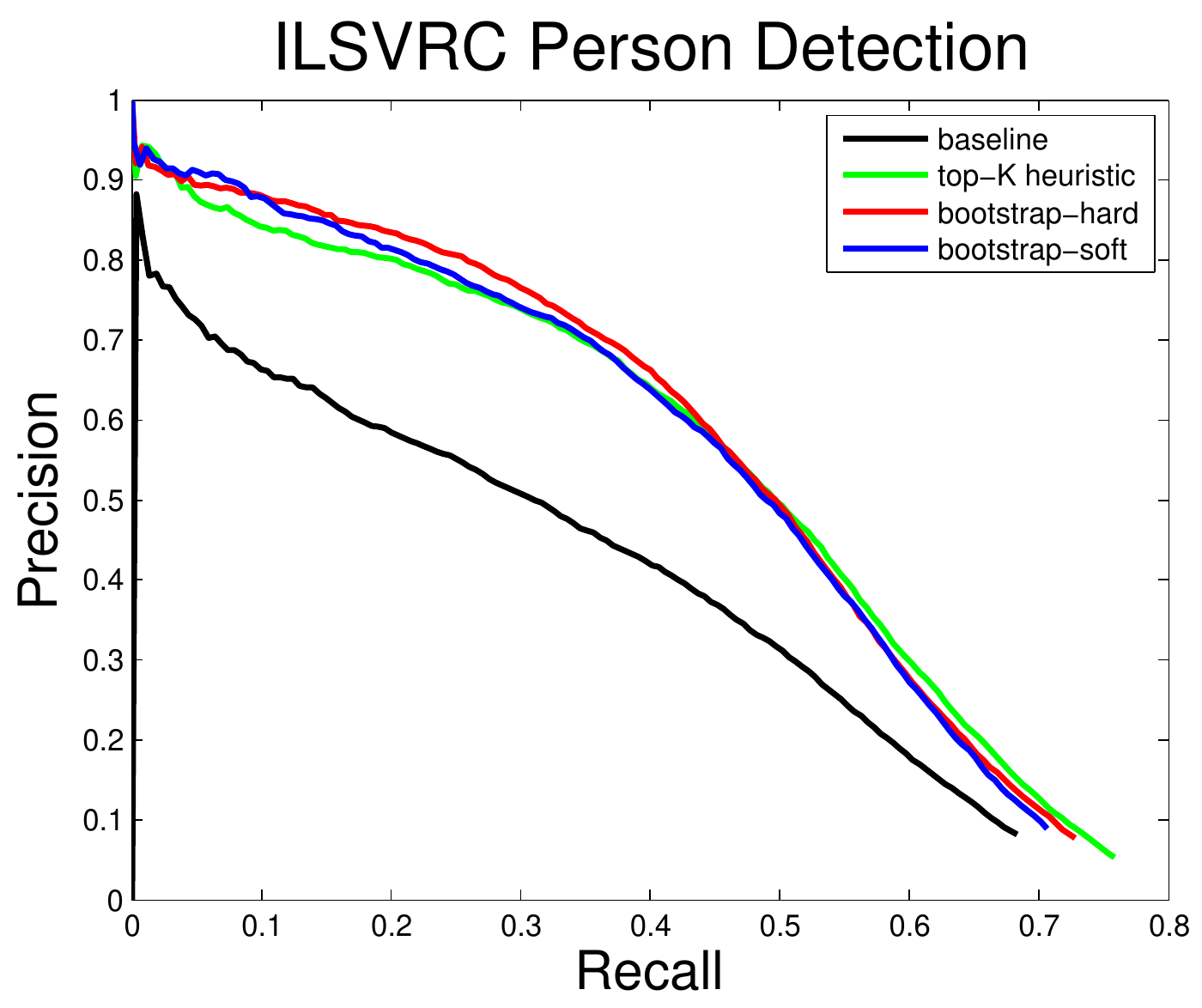}

} {%
  \caption{Precision-recall curves for our methods compared to the baseline prediction-only network, and the top-K heuristic.}%
 \label{fig:person}

}
\killfloatstyle
\btabbox{
  \begin{tabular}{| l | c | c |}
  \hline
  \textbf{Training} & \textbf{AP (\%)} & \textbf{Recall @ 60p} \\
  \hline
  baseline & 30.9 & 14.3 \\
  \hline
  top-K heuristic & 44.1 & 43.4 \\
  \hline
  bootstrap-hard & \textbf{44.6} & \textbf{43.9} \\
  \hline
  bootstrap-soft & 43.6 & 42.8 \\
  \hline
  \end{tabular}
}{%
  \caption{Our proposed bootstrap-hard and bootstrap-soft, and the top-K heuristic significantly improve average precision compared to the prediction-only baseline. Recall@60p is the recall achievable at 60\% precision.}%
}
\end{floatrow}
\end{figure}

The precision-recall curves in figure~\ref{fig:person} show that our proposed bootstrapping improves substantially over the prediction-only baseline. 
%
%
At the high-precision end of the PR curve, the approaches introduced in this paper perform better, while the top-$K$ heuristic is slightly better at high-recall.
%
%
\subsection{ILSVRC 2014 200-category detection}
In this section we apply our method to the case of object detection on the large-scale ImageNet data. 
Our proposed method is applied in two ways: first to the MultiBox network for region proposal, and second to the classifier network that predicts labels for each cropped image region. We follow the approach in~\citep{szegedy2014scalable} and combine image crops from MultiBox region proposals with deep network context features as the input to the classifier for each proposed region.

We trained the MultiBox network as described in~\ref{sec:structured}, and the post-classifier network as described in section~\ref{sec:bootstrap}.
We found that ``hard'' performed better than the ``soft'' form of bootstrapping.

\begin{figure}[h]
  \begin{tabular}{| l | l | c | c |}
  \hline
  \textbf{MultiBox} & \textbf{Postclassifier} & \textbf{mAP (\%)} & \textbf{Recall @ 60p} \\
  \hline
  baseline & baseline & $39.8$  & $38.4$ \\
  \hline
  baseline & bootstrap-hard & $40.0$  & $38.6$ \\
  \hline
  bootstrap-hard & baseline & $40.3$ & $39.3$ \\
  \hline
  bootstrap-hard & bootstrap-hard & $40.3$ & $39.1$ \\
  \hline
  \hline
  - & GoogLeNet single model\footnote{\citep{szegedy2014going}} & 38.8 & -\\
  \hline
  - & DeepID-Net single model\footnote{\citep{ouyang2014deepid}} & 40.1 & -\\
  \hline
  \end{tabular}
\label{fig:ilsvrc_detection}
\end{figure}

Using our proposed bootstrapping method,
%
we observe modest improvement on the ILSVRC2014 detection ``val2'' data\footnote{In a previous version of this draft we included numbers on the full validation set. However, we discovered that context and post-classifier models used ``val1'' data, so we re-ran experiments only on the ``val2'' subset}, mainly attributable to bootstrapping in MultiBox training. 
%
%
\section{Conclusions}
In this paper we developed novel training methods for weakly-supervised deep learning, and demonstrated the effectiveness of our approach on multi-class prediction and structured output prediction for several datasets.
Our method is exceedingly simple and can be applied with very little engineering effort to existing networks trained using a purely-supervised objective.
The improvements that we show even with very simple methods, suggest that moving beyond purely-supervised deep learning is worthy of further research attention.
In addition to achieving better performance with the data we already have, our results suggest that performance gains may be achieved from collecting more data at a cheaper price, since image annotation need not be as exhasutive and mistaken labels are not as harmful to the performance.

In future work, it may be promising to consider learning a time-dependent policy for tuning $\beta$, the scaling factor between prediction and perceptual consistency objectives, and also to extend our approach to the case of a situated agent.
Another promising direction is to augment large-scale training for detection (e.g. ILSVRC) with unlabeled and more weakly-labeled images, to further benefit from our proposed perceptual consistency objective.
\renewcommand*{\bibfont}{\small}
\bibliography{references}
\bibliographystyle{iclr2015}

\end{document}